# Neural Network-Based Equations for Predicting PGA and PGV in Texas, Oklahoma, and Kansas

Farid Khosravikia, M.ASCE[1]; Yasaman Zeinali[2]; Zoltan Nagy[3]; Patricia Clayton[4]; and Ellen M. Rathje[5]

[1] PhD student, Department of Civil, Architectural and Environmental Engineering, The University of Texas at Austin, Austin, TX; e-mail: farid.khosravikia@utexas.edu
[2] Graduate student, Faculty of Civil Engineering, K. N. Toosi University of Technology, Tehran, Iran; e-mail: yas.zeinali@gmail.com
[3] Assistant Professor, Department of Civil, Architectural and Environmental Engineering, The University of Texas at Austin, Austin, TX; e-mail: nagy@utexas.edu
[4] Assistant Professor, Department of Civil, Architectural and Environmental Engineering, The University of Texas at Austin, Austin, TX; e-mail: clayton@utexas.edu
[5] Professor, Department of Civil, Architectural and Environmental Engineering, The University of Texas at Austin, Austin, TX; e-mail: e.rathje@mail.utexas.edu

**ABSTRACT**

Parts of Texas, Oklahoma, and Kansas have experienced increased rates of seismicity in recent years, providing new datasets of earthquake recordings to develop ground motion prediction models for this particular region of the Central and Eastern North America (CENA). This paper outlines a framework for using Artificial Neural Networks (ANNs) to develop attenuation models from the ground motion recordings in this region. While attenuation models exist for the CENA, concerns over the increased rate of seismicity in this region necessitate investigation of ground motions prediction models particular to these states. To do so, an ANN-based framework is proposed to predict peak ground acceleration (PGA) and peak ground velocity (PGV) given magnitude, earthquake source-to-site distance, and shear wave velocity. In this framework, approximately 4,500 ground motions with magnitude greater than 3.0 recorded in these three states (Texas, Oklahoma, and Kansas) since 2005 are considered. Results from this study suggest that existing ground motion prediction models developed for CENA do not accurately predict the ground motion intensity measures for earthquakes in this region, especially for those with low source-to-site distances or on very soft soil conditions. The proposed ANN models provide much more accurate prediction of the ground motion intensity measures at all distances and magnitudes. The proposed ANN models are also converted to relatively simple mathematical equations so that engineers can easily use them to predict the ground motion intensity measures for future events. Finally, through a sensitivity analysis, the contributions of the predictive parameters to the prediction of the considered intensity measures are investigated.

# INTRODUCTION

This paper aims at developing attenuation models for ground motions in Texas, Oklahoma, and Kansas. Attenuation models are generally used to predict earthquake intensity measures given source characteristics of the earthquake, wave propagation path, and local site conditions (Douglas 2003; Kramer 1996). These attenuation models are key in seismic hazard analysis of any region. In this study, peak ground acceleration (PGA) and peak ground velocity (PGV) are the ground motion intensity measures of interest. The attenuation models proposed in this study relate the aforementioned intensity measures to earthquake magnitude, earthquake source-to-site distance, and shear wave velocity of the site.

There has been growing interest in recent decades to develop ground motion prediction equations (GMPEs) for the Central and Eastern North America (CENA) as part of the Next Generation Attenuation (NGA)-East project. One such CENA attenuation models as part of the NGA-East project is developed by Hassani and Atkinson (2015), which is herein referred to as HA15. They employed an empirical approach to develop the ground motion prediction equations for CENA. In this approach, regression analysis is conducted to predict the ground motion intensity measures based on event data from the NGA-East database. However, since 2010, there has been a significant increase in the rate of the ground motions in Texas, Oklahoma, and Kansas (Frohlich et al. 2016; Hornbach et al. 2016; Hough 2014; Petersen et al. 2016). Such studies suggested that much of the recent increase in seismic activity is associated with human activities such as waste fluid injection or extraction. Such activities generally increase the pore pressure, precipitating release of stored tectonic stress along an adjacent fault. Compared to natural tectonic earthquakes, these potentially induced earthquakes are generally shallow-depth events with smaller magnitude, which makes the seismic wave propagation more dependent to the heterogeneous properties of the upper most crustal layers (Bommer et al. 2016). Therefore, it is required to investigate attenuation models for these areas to determine how such earthquake characteristics affect the ground motion prediction equations that have previously been developed based largely on natural tectonic CENA events. These attenuation models can be used to develop seismic hazards for the states, which can be used in seismic risk analysis of different infrastructures in areas with potentially induced seismic hazards (Khosravikia et al. 2017, 2018). In this regard, Zalachoris and Rathje (2017) introduced adjustment factors to the HA15 attenuation models for Texas, Oklahoma, and Kansas, and developed application-specific ground motion prediction equations. Zalachoris and Rathje (2017) showed that the measured response spectral accelerations at short distances for the potentially induced ground motions in these states can be two times larger than spectral accelerations predicted using HA15 attenuation models, which emphasizes the fact that the characteristics of potentially induced earthquakes in those regions require development of attenuation models that are specific to those regions.

This study has three primary objectives, as described below. 1) Attenuation models predicting PGA and PGV for Texas, Oklahoma, and Kansas are developed and compared with existing GMPEs for CENA. Zalachoris and Rathje (2017) updated the HA15 attenuation models for PGA and spectral accelerations at 0.2 and 1 second. The present study not only predicts PGA, but also provides prediction equations for PGV for the ground motions in Texas, Oklahoma, and Kansas. To do so, approximately 4,500 ground motions recorded on Texas, Oklahoma, and Kansas are considered. The characteristics of these ground motions are discussed in the following section. 2) This study proposes an artificial neural network (ANN) based framework for development of the ground motion attenuation models. In conventional empirical methods, generally, regression analysis, using pre-defined linear or nonlinear equations, is conducted to develop attenuation models. This study utilizes an ANN method, which, unlike a regression analysis based on a pre-defined mathematical equation, has the capability of adaptively learning from experience and extracting various discriminators in pattern recognition. This robust method is also used in some related studies to develop ground motion attenuation models for other regions ( Güllü and Erçelebi 2007; Ahmad et al. 2008; Gunaydın and Gunaydın 2008; Alavi and Gandomi 2011;). One of the main drawbacks of ANN method is that it is often used as a black-box system that it is not able to present the underlying principles of the prediction. To solve this problem, the ANN models are here converted to relatively simple mathematical equations. The ANN-based framework proposed in this study is discussed in detail in the following sections. 3) Through a sensitivity analysis, the importance of each predictive variable is computed and compared for each intensity measure.

**GROUND MOTION RECORDED ON TEXAS, OKLAHOMA, AND KANSAS**

This study takes into account 4,529 ground motion recordings with epicenters in Texas, Oklahoma, and Kansas. These ground motions were retrieved from the Incorporated Research Institutions for Research, IRIS database (https://www.iris.edu/hq/), and processed by Zalachoris and Rathje (2017). These ground motion recordings correspond to 374 different earthquake events. It should also be noted that the ground motions recorded at seismic stations on the Gulf Coast Plain are not considered in this study because the significantly different geologic characteristics of the Gulf Coast Plain result in different site amplification for this region (Zalachoris and Rathje 2017). All of the selected events have moment magnitudes greater than 3.0 and occurred after 2005. Figure 1 demonstrates the frequency of the main seismic characteristics of these events including: moment magnitude, $M_w$, shear wave velocity in the upper 30m of soil, $V_{s30}$, Joyner-Boore distance, $R_{JB}$, as well as the natural log of peak ground acceleration, PGA, and peak ground velocity, PGV.

The magnitudes of these ground motions are either given in IRIS or computed using 1-Hz PSA amplitudes of the vertical component of the ground motion records (Atkinson and Mahani 2013; Atkinson et al. 2014). As seen in the figure, the considered events cover the magnitude range between 3 and 5.8, which represents small to moderate earthquakes. The earthquake source-to-site distance, here, is presented by Joyner-Boore distance, $R_{JB}$, which is defined as the closest distance to the surface projection of the rupture. For the events considered in this study, $R_{JB}$ is approximately equal to the epicentral distance (Hassani and Atkinson 2015). Figure 1 shows that $R_{JB}$ varies between 4 km to 500 km. It should be noted that approximately 852 records, or 18.8%, have $R_{JB}$ less than 50 km.

In the developed ANN model, the averaged shear wave velocity over the top 30 m of soil, $V_{s30}$, is used as a measure of the site amplification due to local site conditions. The values of $V_{s30}$ for Texas, Oklahoma, and Kansas are provided by Zalachoris et al. (2017). They used the P-wave seismogram method to estimate $V_{s30}$ at numerous seismic stations over these states. For a few seismic station locations that P-wave seismogram $V_{s30}$ estimates were not available, the $V_{s30}$ estimates provided by Parker et al. (2017) are utilized. Parker et al. (2017) predicted the values of $V_{s30}$ over the CENA using a hybrid slope-geology proxy method. It is estimated that 92 of the 209 considered seismic stations have $V_{s30}$ greater than 760 m/s, which refers to "rock" site conditions. According to National Earthquake Hazards Reduction Program (NEHRP) site classification, these seismic stations are classified as Site Classes A and B (FEMA 2015).

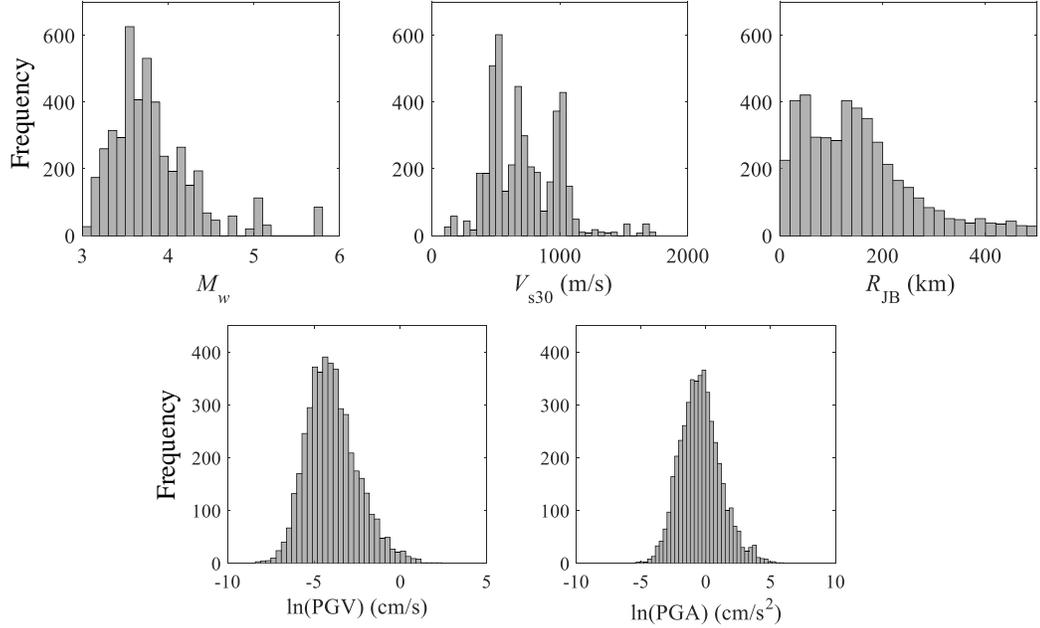

Figure 1: Histogram of the ground motion characteristics in Texas, Oklahoma, and Kansas

**NEURAL NETWROK BASED FRAMEWORK**

Artificial neural network (ANN), founded by McCulloch and co-workers in the early 1940s (Perlovsky 2001), is a statistical learning model, inspired by biological neural networks in the human brain. ANN can be used in different problems without algorithmic solutions or problems with complex solutions to provide a mathematical model to predict the outputs given the inputs. In this study, a multilayer perceptron (MLP) network, which is a kind of ANN that has a feed-forward architecture, is used to predict PGA and PGV (Cybenko 1989). MLP network consists of three different layers, namely input, hidden, and output layers. The neurons of each layer are fully connected to the neurons of other layers with connections weights. The connection weights come from the training process of the network, which is done here by implementing the Levenberg–Marquardt back-propagation algorithm (Marquardt 1963). To train the ANN in this study, the 4,529 ground motion recordings are randomly divided into three different subsets: training, validation, and testing subsets, which respectively consist of 60%, 20%, and 20% of the whole data set. Training and validation sets are used for training the ANN. In particular, the training subset is used to adjust the weights and bias values on the ANN model, and the validation subset is used to minimize overfitting of the model by checking the generalization capability of the models on data they did not train on. Generally, during the training process, the accuracy over the training subset increases. If the accuracy over the validation subset stays the same or decreases with increased training, then an overfitting problem occurs, and the training process is stopped. Finally, the testing subset is used for testing the final algorithm to confirm the actual predictive power of the network for future data.

In this study, two different ANN models are developed to separately predict PGA and PGV given the values of earthquake magnitude, $M_w$, site shear wave velocity, $V_{s30}$, and Joyner-Boore distance, $R_{JB}$. Figure 2 presents the schematic view of the proposed ANN framework for PGA. As seen, the input layer consists of three neurons representing normalized values of $M_w$, $V_{s30}$, and $R_{JB}$, and the output layer comprises one neuron representing normalized values of PGA. Normalizing the inputs and the outputs in ANN generally provides better accuracy for the developed models. In this algorithm, one hidden layer with four neurons is considered. There are two criteria in determining the number of neurons for the hidden layer as follows: The number of neurons in the hidden layer must 1) lead to the simplest network, which requires fewest number of neurons, while 2) providing sufficient accuracy for training and test data sets. The accuracy here is conducted by *R*-square analysis. By trainings different networks with of the number of neurons for the hidden layer varying from 1 to 10, it was concluded that the ANN model with four neurons in the hidden layer is the simplest network that results in very good accuracy, as will be discussed in later sections.

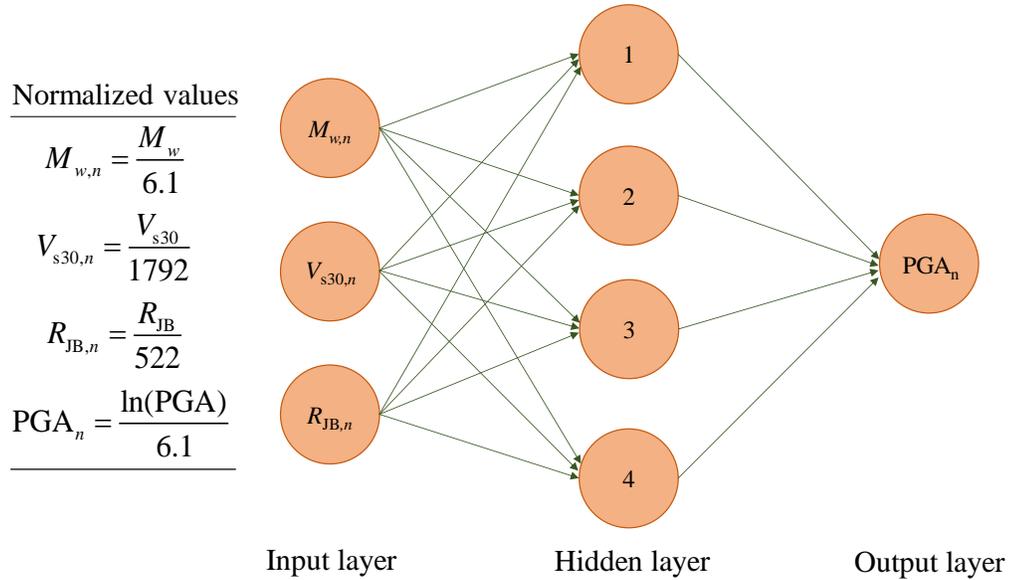

Figure 2: Schematic representation of ANN model for PGA

Figure 3a shows the general structure of *i*-th neuron from the hidden layer. As seen, each neuron receives inputs that are weighted with corresponding connection weight. The summation of the weighted inputs and the bias, $b_i$, forms the input to the activation function. Passing the summation through the activation function, the output of the neuron is computed using the following equation:

$$y_i = \varphi(w_{1i} \times M_{w,n} + w_{2i} \times V_{s30,n} + w_{3i} \times R_{JB,n} + b_i) \tag{1}$$

where $y_i$ is the output of the *i*-th neuron in the hidden layer; $w_{ji}$ is the connection weight of the *j*-th neuron from the input layers and *i*-th neuron from the hidden layer; $b_i$ is the bias defined for *i*-th neuron in this hidden layer; and $\varphi$ is the activation function for the neurons in the hidden layer, which is here assumed to follow a log-sigmoid function of $\varphi(x) = (\frac{1}{1+e^{-x}})$. Other types of activation function can also be used for ANN network. In fact, the accuracy of the ANN model does not necessarily depend on the selection of the activation function. However, it affects the values of the weights between different nodes. The bias parameter acts like an input neuron that always has the value of 1.0 with connection weight of $b_i$. The value of $b_i$ is determined during the training process. Bias parameter, which provides considerable flexibility to an ANN model, is analogous to the intercept in a regression model. In fact, when the input values are zero, the summing junction can only produce zero unless it has the bias parameter. The output, $y_i$, will be, in turn, the input of the neurons in the next layer, i.e. neurons in the output layer, to predict the output parameters. Figure 3b shows the general structure of output neuron. As seen, it follows the same structure as the neurons in the hidden layer. The values of $y_i$ computed in the hidden neurons are set as inputs for this neuron. Passing through the weighted summation from the activation function, the values of PGA are predicted. To do so, a linear activation function is adopted for the neurons of the output layer. The linear activation function scales the output of the summation junction to the actual values of the ground motion parameters. The same framework is used for predicting PGV. The only difference between ANN models of PGA and PGV, here, is the connections weights, which are separately computed through the training process.

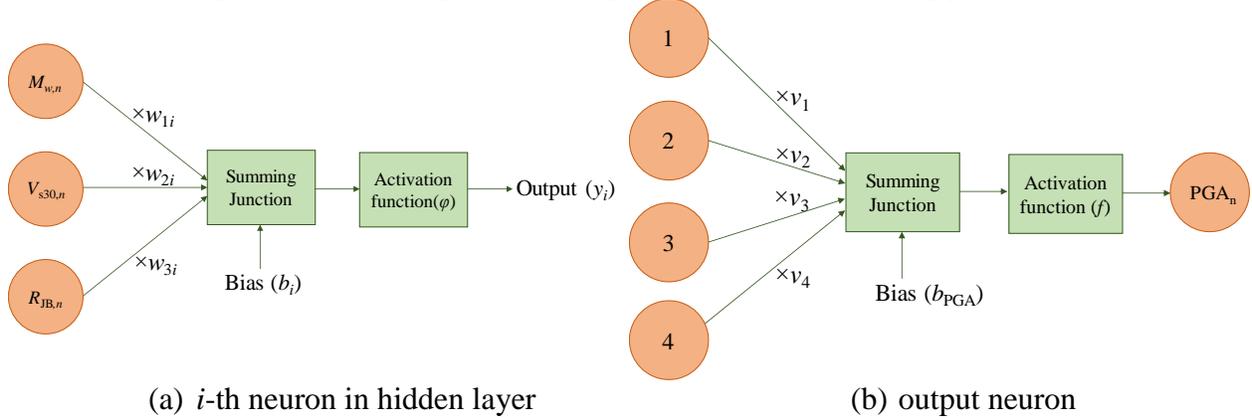

(a) $i$-th neuron in hidden layer  (b) output neuron

Figure 3: Schematic representation of neurons in the proposed ANN model for PGA

**ATTENUATION MODELS**

ANN models discussed in the previous section are trained and turned into the mathematical formulations of PGA (cm/s$^2$) and PGV (cm/s), as follows:

$$\begin{pmatrix} \dfrac{\ln(\text{PGA})}{6.1} \\ \dfrac{\ln(\text{PGV})}{2.5} \end{pmatrix} = \begin{pmatrix} b_{\text{PGA}} + \sum_{i=1}^{4} v_{i,\text{PGA}} \times \left(\dfrac{1}{1+e^{-F_{i,\text{PGA}}}}\right) \\ b_{\text{PGV}} + \sum_{i=1}^{4} v_{i,\text{PGV}} \times \left(\dfrac{1}{1+e^{-F_{i,\text{PGV}}}}\right) \end{pmatrix} \quad (2)$$

where $b_{\text{PGA}}$ and $b_{\text{PGV}}$ represent the bias values of output neurons for ANN models of PGA and PGV, respectively; $v_{i,\text{PGA}}$ and $v_{i,\text{PGV}}$, denotes the connection weights between the $i$-th neuron from the hidden layer and output neuron for each ANN model; $F_{i,\text{PGA}}$ and $F_{i,\text{PGV}}$ are computed using the following equations:

$$\begin{pmatrix} F_{i,\text{PGA}} \\ F_{i,\text{PGV}} \end{pmatrix} = \begin{pmatrix} w_{1i,\text{PGA}} \times \dfrac{M_w}{6} + w_{2i,\text{PGA}} \times \dfrac{V_{s30}}{1792} + w_{3i,\text{PGA}} \times \dfrac{R_{\text{JB}}}{522} + b_{i,\text{PGA}} \\ w_{1i,\text{PGV}} \times \dfrac{M_w}{6} + w_{2i,\text{PGV}} \times \dfrac{V_{s30}}{1792} + w_{3i,\text{PGV}} \times \dfrac{R_{\text{JB}}}{522} + b_{i,\text{PGV}} \end{pmatrix} \quad (3)$$

where $w_{ji,\text{PGA}}$ and $w_{ji,\text{PGV}}$ respectively refer to the connection weights between the $j$-th neuron from the input layer and $i$-th neuron from the hidden layer of ANN models for PGA and PGV. $b_{i,\text{PGA}}$ and $b_{i,\text{PGV}}$ denote the bias of the hidden layer neurons for each ANN model. Table 1 shows the connection weight and bias values of the proposed ANN models after the training process using the Levenberg–Marquardt algorithm (Marquardt 1963).

Table 1: Connection weights and bias values for the proposed ANN models

| PGA model | | | | | | |
|---|---|---|---|---|---|---|
| Hidden neuron($i$) | $w_{1i,\text{PGA}}$ | $w_{2i,\text{PGA}}$ | $w_{3i,\text{PGA}}$ | $b_{i,\text{PGA}}$ | $v_{i,\text{PGA}}$ | $b_{\text{PGA}}$ |
| 1 | -93.7502 | -0.1658 | -4.7160 | 68.6111 | -0.1037 | |
| 2 | 4.9023 | -0.6769 | -2.7333 | -2.6134 | 1.1886 | -0.6149 |
| 3 | -1.3182 | 0.9545 | -43.7438 | -1.4151 | 6.5491 | |
| 4 | 21.7529 | 2.5431 | -6.6562 | -9.8652 | 0.1886 | |
| PGV model | | | | | | |
| Hidden neuron($i$) | $w_{1i,\text{PGV}}$ | $w_{2i,\text{PGV}}$ | $w_{3i,\text{PGV}}$ | $b_{i,\text{PGV}}$ | $v_{i,\text{PGV}}$ | $b_{\text{PGV}}$ |
| 1 | 1.7409 | -0.4457 | 45.7174 | 1.1633 | -15.1236 | |
| 2 | -2.0083 | 0.0730 | 0.2576 | 0.3429 | -12.4700 | 18.0142 |
| 3 | -0.9230 | 0.6639 | 10.4003 | -1.7592 | -2.6548 | |
| 4 | -2.3723 | -0.5214 | 18.8468 | -2.6345 | 1.6283 | |

Figure 4 shows the correlation between the predicted and measured values for the ANN models. Each column of plots presents the results for one of the developed ANN models, e.g. PGA or PGV. The top and bottom plots in each column, respectively, present the results for training and testing subsets. The large values of the correlation coefficient, $R$, demonstrates that the models provide acceptable estimates of the target values. Moreover, having large values of $R$ for both training and test subsets shows that the models have very good ability of prediction and generalization performance; hence, they can reliably be used to determine the principal ground-motion parameters. It is worth noting that the predication reliability is limited for ground motions with the same range of the input characteristics shown in Figure 1.

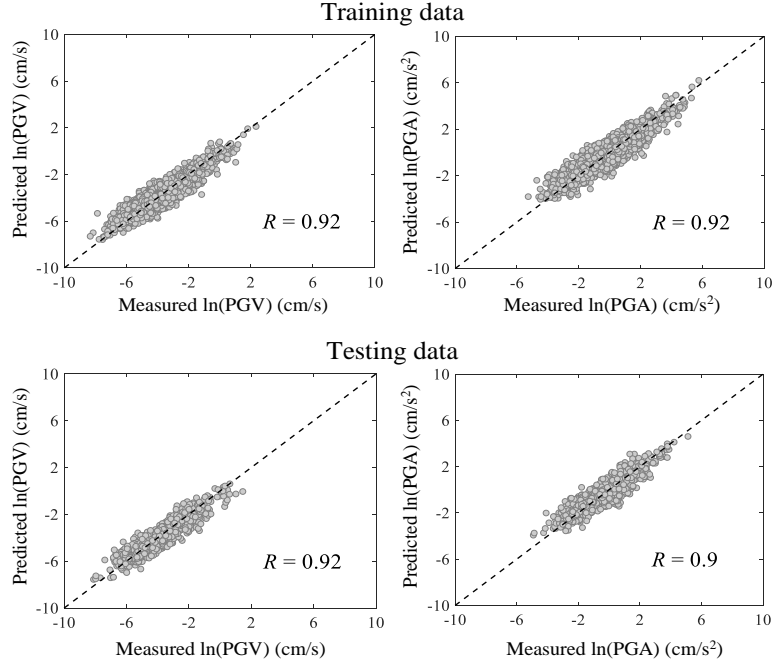

Figure 4: Measured versus predicted values from ANN models

## ATTENUATION MODEL COMPARISON

Recall that Hassani and Atkinson (2015) developed ground motions prediction equations for CENA, which are referred to as HA15 in this study. It was also mentioned that the characteristics of potentially induced ground motions in Texas, Oklahoma, and Kansas may be different from natural earthquakes in the CENA. In particular, they are generally shallower with lower magnitudes compared to natural seismic hazards in CENA. One way to investigate how such characteristics may result in disparities between the predicted and measured ground motion parameters is to plot residuals versus ground motion parameters. residuals are, here, defined as the natural log of the ratio of the measured over predicted ground motion parameters (Boore and Atkinson 2007). Here, for brevity, the results are only shown for the PGA attenuation model. The right two plots of Figure 5 present the residuals from HA15 in relation to distance as well as shear wave velocity for the ground motion database used in this study. The positive values of residual for $R_{JB} < 20$ km in the left figure shows that HA15 underestimates the values of PGA for ground motions that have occurred in Texas, Oklahoma, and Kansas. For instance, the mean residuals for HA15 are as large as 0.8 for $R_{JB} < 20$ km, which implies that the observed values of PGA are nearly 2.2 times larger than the predicted values. This trend is likely owing to the fact that the potentially induced seismic hazards in Texas, Oklahoma, and Kansas have shallower depth compared to the natural CENA earthquakes that served as the basis of HA15. For higher values of $R_{JB}$, the negative values of residual demonstrate that HA15 overestimates the PGA values for ground motions at longer distances from the epicenter. This observation implies that the ground motions at longer distances in Texas, Oklahoma, and Kansas are not as intense as what is expected in CENA.

The residuals are also plotted versus $V_{s30}$, which represents site effects. The predominantly negative residuals on the left figure, especially for lower values of $V_{s30}$, indicates that PGA of the observed ground motions are smaller than what is predicted by HA15. This overprediction is owing to the fact that the models developed for CENA use an amplification model originally developed for a Western U.S., which is characterized by deeper sediments and results in larger amplifications (Zalachoris et al. 2017). However, most of the sites in Texas, Oklahoma, and Kansas with very soft soils tend to have low depth to bedrock (Zalachoris et al. 2017). The results in the figure are presented with different types of dots based on the range of the PGA. As seen, most of the data are associated with small values of PGA, i.e. less than 0.05 g; therefore, the residuals for the Hassani Atkinson (2015) model with respect to $V_{s30}$ are representative of the linear-elastic part of the site amplification model.

The residuals for the developed ANN model are also shown in Figure 5. As seen, the average of the residuals is zero, regardless of the values of $R_{JB}$ and $V_{s30}$, which means the ANN model on average accurately predicts the PGA values for the considered ground motions. This improved accuracy in the ANN model does not raise any question about the accuracy of the attenuation models developed by Hassani and Atkinson (2015) as their proposed prediction models for the CENA were developed using another database of events from all over the CENA. This figure shows that if one wants to use their prediction models for Texas, Oklahoma, Kansas, he or she will probably get inaccurate results. However, the ANN models result in much more accurate site-specific results in this particular application.

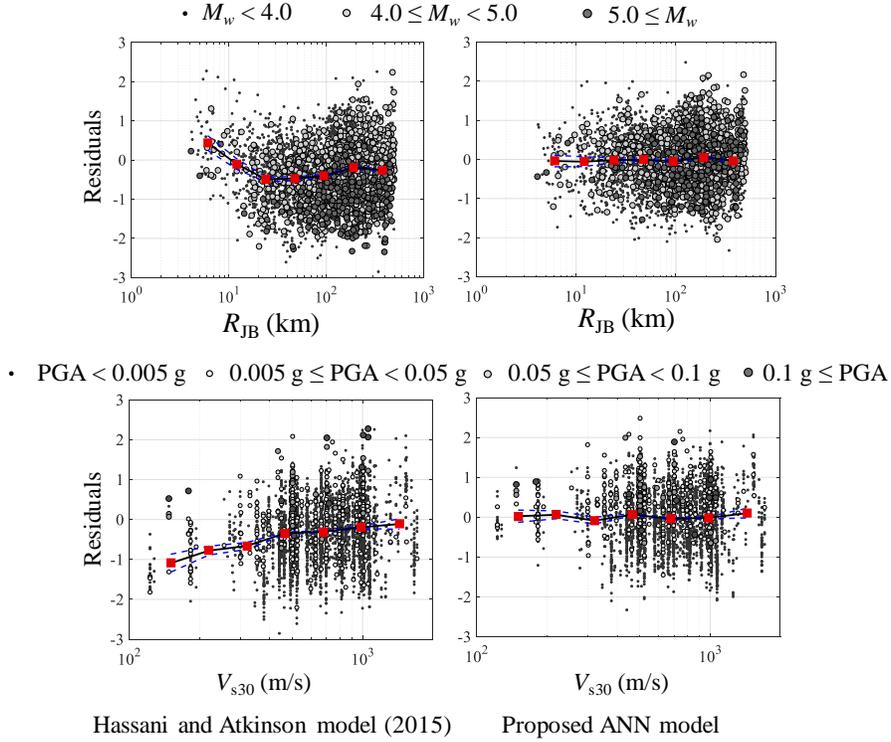

Note: The mean and 90% confidence interval values are respectively shown by red and blue curves.

Figure 5: Residuals of the models versus input characteristics

Finally, Figure 6 presents the PGA distance relations for the developed ANN models. For purposes of demonstration, the results are shown separately for ground motions with magnitude range of larger than 5.0, and for ground motions with magnitude range between 3.5 to 4.0. The curves correspond to the magnitudes of 3.7 and 5.3 as well as $V_{s30}$ equal to 760 m/s. The solid and dashed curves show the results from ANN model proposed in the present study and HA15 attenuation models, respectively.

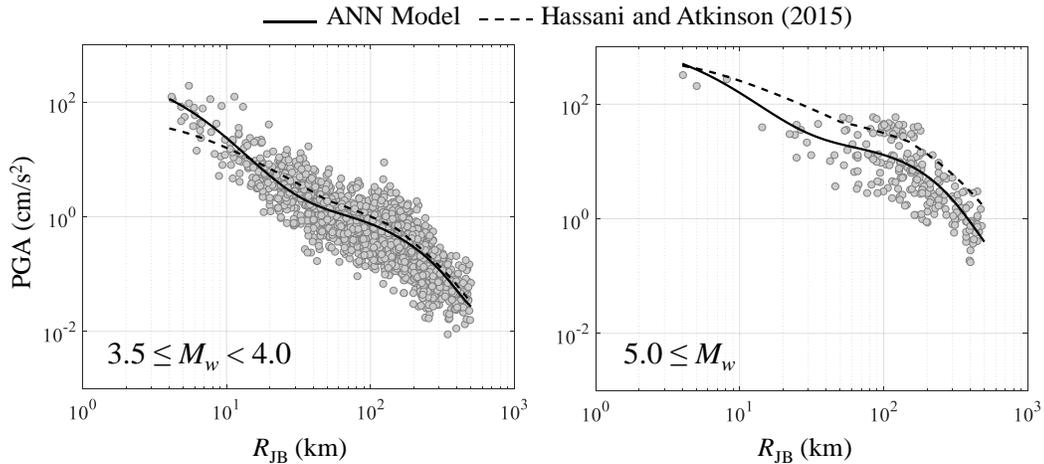

Note: solid and dashed curves in the left plot are depicted for $V_{s30}$ = 760 m/s and $M_w$ = 3.7, and in right plot are depicted for $V_{s30}$ = 760 m/s and $M_w$ = 5.3.

Figure 6: The PGA to distance relations for different magnitude ranges

As seen in Figure 6, for ground motions with lower magnitudes, HA15 underestimates the PGA values at short distances. Furthermore, for higher magnitudes or longer distances, HA15 overestimates the PGA values. However, the proposed ANN model for PGA matches the data reasonably well for both magnitude ranges at all distances.

**SENSITIVITY ANALYSIS**

In this section the sensitivity of the ANN attenuation models to the predictor variables are evaluated. Recall that the predictor variables in this study are moment magnitude, $M_w$, shear wave velocity, $V_{s30}$, and Joyner-Boore distance, $R_{JB}$. Garson's algorithm (Garson 1991) is used to compute the contribution of each input variable in the ANN output. In this algorithm, the input-hidden and hidden-output weights of the trained ANN models are partitioned, and the absolute values of the weights are taken to calculate the relative importance values. The computed importance values for each ANN model are shown in Figure 7. As seen, for PGA, moment magnitude of the earthquake is the most important parameter, while for PGV, source-to-site parameter is the most important one. For both intensity measures, $V_{s30}$, which represents the site condition, has relatively low contribution.

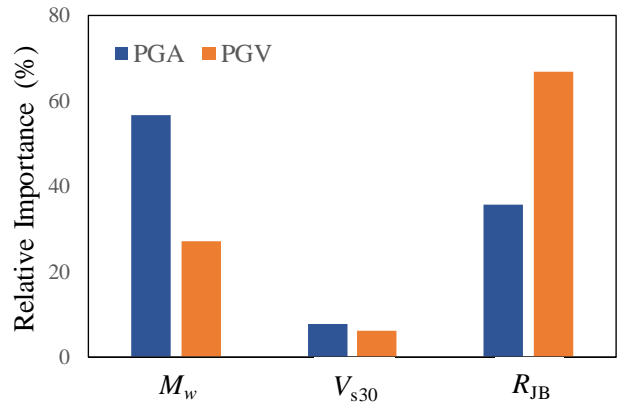

Figure 7: Contribution of the predictor variables in the ANN models of PGA and PGV

**CONCLUSION**

This paper proposes ground motion prediction models developed using Artificial Neural Network (ANN) methods for states of Texas, Oklahoma, and Kansas. These states are believed to be recently subjected to induced seismic hazards as a consequence of human activities associated with petroleum activities and waste water disposal. Compared to natural seismic hazards, most of the induced earthquakes are shallow-depth earthquake with low magnitude. To investigate potential differences between these two types of seismic hazards, ground motion predictions models for those areas that are subjected to the potential human-caused earthquakes were developed and compared to existing CENA models based on natural tectonic events. Here, two fundamental ground motion parameters, namely peak ground acceleration (PGA) and peak ground velocity (PGV) are predicted based on earthquake magnitude, earthquake source-to-site distance, and average shear wave velocity over the top 30m of soil. In the literature, regression analysis is the most common technique to develop the attenuation models; however, using pre-defined linear or nonlinear equation limits the attenuation models in their ability to efficiently simulate the complex behavior of the ground motions characteristics. This study uses ANN methods, which is one of the state-of-the-art machine-learning techniques, to develop the attenuation models. The ANN method is able to adaptively learn from experience and extract various discriminators in pattern recognition.

The results show that the existing CENA ground motion prediction models are not able to properly predict the intensity measures of the ground motion recorded in Texas, Oklahoma, and Kansas. In particular, they are likely to result in underprediction for ground motions at short distances and overprediction for ground motions recorded on soft soils. Moreover, it is verified that the proposed ANN models can accurately predict the values of PGA and PGV in these states. The proposed ANN models are then turned into mathematical equations so that engineers can easily use them. Finally, through a sensitivity analysis, it is revealed that for potentially induced ground motions in Texas, Oklahoma, and Kansas, magnitude and source-to-site distance are the most dominant predictive parameter for PGA and PGV, respectively. On the other hand, it is also shown that, shear wave velocity has low influence on determining the PGA and PGV values. Finally, it is worth noting that the predication reliability and accuracy of the sensitivity analysis are limited for ground motions within the similar range of the input characteristics considered in this study.

**AKNOWLEDGEMENT**

The financial support from the Texas Department of Transportation (TxDOT) through Grant Number 0-6916 is gratefully acknowledged. The opinions and findings expressed herein are those of the authors and not the sponsors.